\documentclass{article}
\usepackage[nonatbib]{neurips_2019}
\usepackage[utf8]{inputenc} 
\usepackage[T1]{fontenc}    
\usepackage{url}            
\usepackage{booktabs}       
\usepackage{amsfonts}       
\usepackage{nicefrac}       
\usepackage{microtype}      
\usepackage{graphicx}
\usepackage{xcolor}
\usepackage{subcaption}

\title{Zero-shot Active Learning Using Self Supervised Learning\\
}

\begin{document}
\maketitle
\section{Introduction}
Deep learning algorithms are often said to be data hungry. The performance of such algorithms generally improve as more and more annotated data is fed into the model. While collecting unlabelled data is easier (as they can be scraped easily from the internet), annotating them is a tedious and expensive task. Given a fixed budget available for data annotation, Active Learning helps selecting the best subset of data for annotation, such that the deep learning model when trained over that subset will have maximum generalization performance under this budget. \\

A common theme in active learning is to start with a random subset of labelled data and then use different strategies to iteratively add new datapoints. We believe that there are three limitations of the current active learning approaches. 
\begin{itemize}
    \item For the current approaches examples are added iteratively in batches, which can make this process difficult to scale. If an orgnization uses a third party system to do data annotation process, then back and forth iterative communication between them can make the process tedious and difficult to use.
    \item Active learning approaches are not model agnostic. It means that for any new model, another round of Active learning will have to be run to get the best subset of data given the budget constraints.
    \item The current approaches start with a random subset of data. Thus the initial data that is collected for labelling is not an optimal subset.
\end{itemize}
In this work, we aim to propose a new Active Learning approach which is model agnostic as well as one doesn't require an iterative process. We aim to leverage self-supervised learnt features for the task of Active Learning. The benefit of self-supervised learning, is that one can get useful feature representation of the input data, without having any annotation. 

\section{Related Work}
\subsection{Active Learning}
 Two popular directions for Active Learning are -  
\begin{itemize}
    \item Uncertainty Estimation - ~\cite{gal2017deep, yoo2019learning, 10.1145/3308560.3316599} Select datapoints for which the model is most uncertain about. The idea here is that adding such uncertain data points will help the model better learn about the data, which in turn help it to generalize well.
    \item Diversity - ~\cite{sener2017active, kirsch2019batchbald, singh2019one} - Select datapoints such that the diversity of labelled set is increased. ~\cite{sener2017active, singh2018footwear} uses core-set/k-centre greedy approach to iteratively sample new datapoints, using features extracted from the model that has been trained over the current pool of labelled data. We will base our algorithm over this approach.
\end{itemize}

\subsection{Self-Supervised Learning}

Self-supervised learning ~\cite{chen2020big, chen2020improved, kumari2017parallelization}, is another area of research which aims to do away with the expensive data annotation process. The idea here is to learn good features from the data, which can then be used for other downstream tasks. 

\begin{figure}
    \centering
    \includegraphics[width=0.65\textwidth]{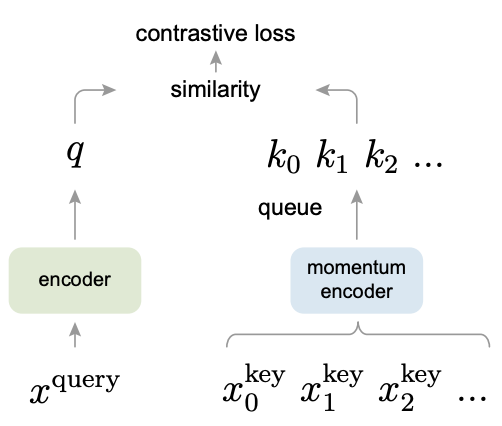}
    \caption{Schematic Overview of Momentum Contrastive Learning Method}
    \label{fig:moco}
\end{figure}

\paragraph{Momentum Contrastive Learning~\cite{chen2020simple, 10.1007/978-981-10-8639-7_25}.} Momentum Contrast (MoCo) trains a visual representation encoder by matching an encoded query $q$ to a dictionary of encoded keys using a contrastive loss. The dictionary keys $\{k_0, k_1, k_2, ...\}$ are defined on-the-fly by a set of data samples. The dictionary is built as a queue, with the current mini-batch enqueued and the oldest mini-batch dequeued, decoupling it from the mini-batch size. The keys are encoded by a slowly progressing encoder, driven by a momentum update with the query encoder. The learned representations are then evaluated by training a linear classifier using labeled training data and evaluating on a test data. This method enables a large and consistent dictionary for learning visual representations. 
Figure \ref{fig:moco} shows a schematic overview of MoCo.

\section{Datasets}
We will be performing our experiments over CIFAR-10 dataset~\cite{krizhevsky2009learning, jindal2023classification}.
The CIFAR-10 dataset consists of 60000 32x32 colour images in 10 classes, with 6000 images per class. There are 50000 training images and 10000 test images. The dataset is publicly available. 

\section{Method}
We aim to use self-supervised for the task of active learning. Since the features can be extracted without any annotation, these features are agnostic to the model as well as the task which will be used for a given dataset. Further, since these features are pre-computed once, and do not change as a downstream task is performed for the dataset, we can get an ordering in which the data needs to be annotated from before without going into an iterative process. Specifically, our approach consists of the following steps - 

\begin{itemize}
    \item Train a self-supervised model over the CIFAR-10 dataset, or use an existsing self-supervised model trained over the ImageNet dataset~\cite{deng2009imagenet}.
    \item Get features for all the images in the CIFAR-10 dataset.
    \item Apply k-centre greedy algorithm to get the ordering in which the dataset needs to be labelled~\cite{sener2017active, rastogi2023exploring}.
    \item Given the budget constraint select the top ordered data-points, annotate them and then train a model for this subset.
\end{itemize}
In the k-centre greedy approach we start with $k$ randomly selected datapoints as cluster centres. Given a fixed budget $B$ we iteratively add $B$ datapoints by selecting the one which is farthest away from the cluster centres. Once a datapoint is selected, it is also added as a new cluster centre. This is a diversity based Active Learning approach, as we want the selected data-points to be as diverse as possible\\
\section{Experiment Details}
We perform our experiments over the CIFAR-10 dataset, and the task is classification. We choose Wide ResNet~\cite{zagoruyko2016wide} as the model for this task. We further use CutOut~\cite{devries2017improved} augmentation to improve the generalization ability.\\

We evaluate our approach for different budget constraints, where the budget is defined by the number of annotated/labelled data-points. We show results for budgets of 5k to 25k with an interval of 2500. Thus if the budget is $B$, we select $B$ data-points, annotate them, train our Wide ResNet using the $B$ data-points and then report the test accuracy of this model.\\

\subsection{Baselines} 
We choose the following two baseline approaches  for comparison - 
\begin{itemize}
    \item \textbf{Random} - For each of the budget contraint, randomly select those number of datapoints from the dataset.
    \item \textbf{Core-Set}~\cite{sener2017active} - Same as our approach, except that it uses a supervised model instead of self-supervised model. Thus, given a subset of annotated datapoints till now, it trains a supervised model over it, and then uses the pre-final layer embeddings as the features of an input. The features are used to perform k-centre greedy approach and then select the new data-points. Since, for any selected subset we need to train a model to get the features, thus the features will be dynamically updated for every iteration of Active learning. This is in contrast to us, where our features are fixed throughout the Active Learning process \cite{rastogi2023exploring}.
\end{itemize}

\subsection{Active Learning Using Self-Supervised Features (Our Approach)}
We  use MoCo-v2~\cite{chen2020improved} and SimCLR~\cite{chen2020simple, rajan2023shaping}  to perform self-supervised learning over the CIFAR-10 dataset. 
We use the following data-augmentation strategies for MoCo-v2 training - 
\begin{itemize}
    \item \textbf{Random Grayscale} - Images are randomly converted to grayscale with a probability of 0.2
    \item \textbf{Random Resize Crop} - A crop of scale randomly chosen between 0.08 to 1.0  is made, before resizing the crop again to size 32*32.
    \item \textbf{Random Horzintal Flip} - The image is randomly flipped horizontally with a probability pf 0.5.
    \item \textbf{Random Color jitter} - The brightness, hue and saturation are randomly changed with a proabibility of 0.8.
\end{itemize}
To test the efficacy of the self supervised learning, we use the linear evaluation protocol ~\cite{chen2020improved, chen2020simple} where the network is frozen, and only a linear layer is learnt over the labelled trainset. 
On the linear evaluation protocol, MoCo-v2 achieves a test accuracy of 91.2\% whereas SimCLR achieves a test accuracy of 93.6\% 

\subsection{Evaluation}
As mentioned above we train different WideResNet models for each of the budget constraints, and then compare the test accuracies of the different algorithms for each budget.

\subsection{Results}
\begin{figure}[!ht]
    \centering
    \includegraphics[width=\linewidth]{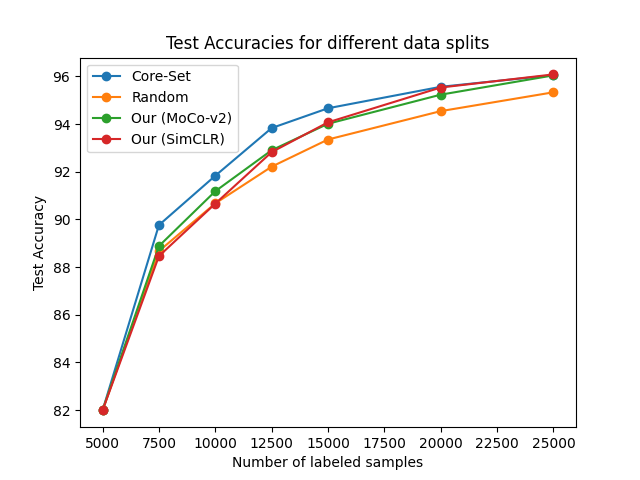}
    \caption{Performance of different approaches}
    \label{fig:our}
\end{figure}

Figure ~\ref{fig:our} shows the results for the two baselines as well as our approach over different data splits.
 As can be seen from the figure, Our approach performs better than the random baseline, but performs slighly worse the Core-Set baseline when the number of labelled samples are less than 20k. For data split sizes of 20k and more, our approach is comparable to the Core-Set baseline. \\
 
 Our approach although performs comparable or slightly worse than the Core-Set baseline, it has the advantage that the method is non-iterative based and also is model agnostic. Thus, even if instead of Wide ResNet we used any other model for classification, the same sampled data-points could have been used for labelling.

\section{Discussion}

\begin{figure}
     \centering
     \begin{subfigure}[b]{0.45\textwidth}
         \centering
         \includegraphics[width=\textwidth]{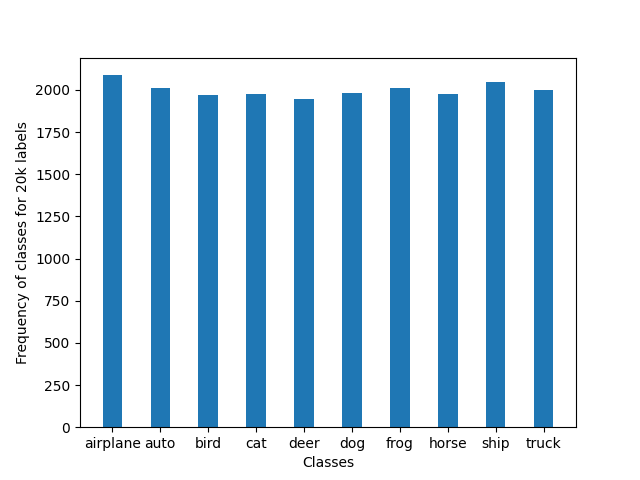}
         \caption{Random}
         \label{fig:random_bar}
     \end{subfigure}
     ~
     \begin{subfigure}[b]{0.45\textwidth}
         \centering
         \includegraphics[width=\textwidth]{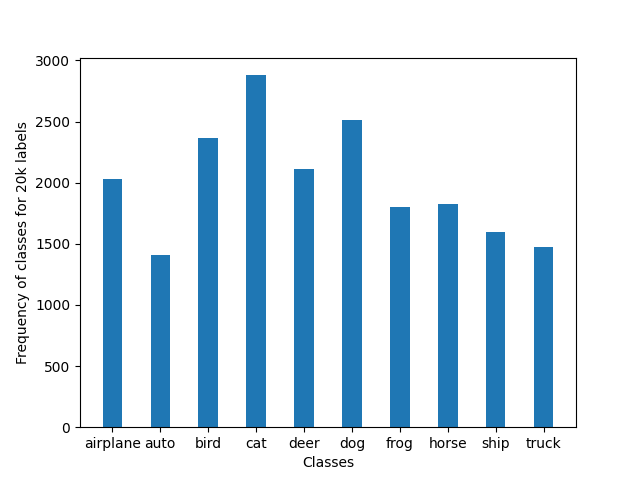}
         \caption{Core-Set}
         \label{fig:core-set_bar}
     \end{subfigure}
    \caption{Frequency of classes in the 20k data points selected for labelling for the Baselines}
    \label{fig:baseline_bar}
\end{figure}

\begin{figure}
     \centering
     \begin{subfigure}[b]{0.45\textwidth}
         \centering
         \includegraphics[width=\textwidth]{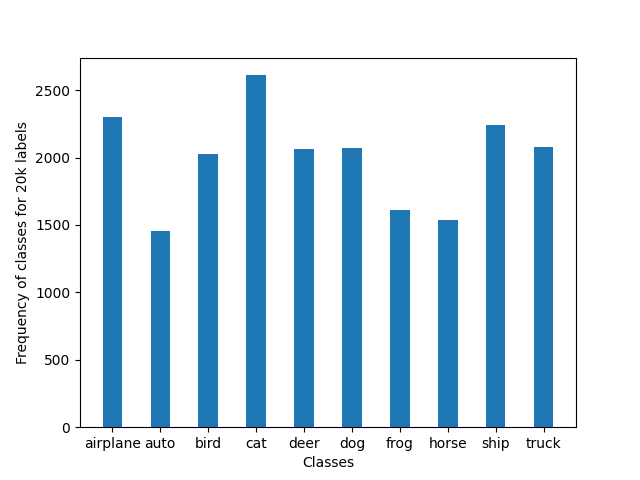}
         \caption{Our (MoCo-v2)}
         \label{fig:our1_bar}
     \end{subfigure}
     ~
     \begin{subfigure}[b]{0.45\textwidth}
         \centering
         \includegraphics[width=\textwidth]{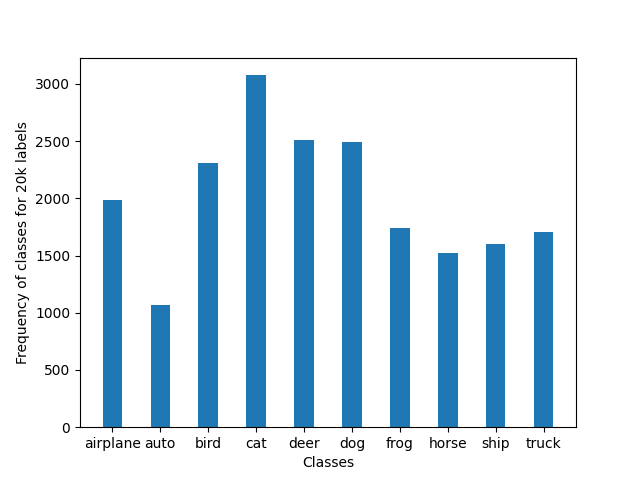}
         \caption{Our (SimCLR)}
         \label{fig:our2_bar}
     \end{subfigure}
    \caption{Frequency of classes in the 20k data points selected for labelling for the Baselines}
    \label{fig:our_bar}
\end{figure}

We plot the frequency of each of the classes amongst the 20k data points selected  for labelling in Figures ~\ref{fig:baseline_bar} and ~\ref{fig:our_bar}.\\

As can be seen in Figure ~\ref{fig:random_bar} random selection uniformly selects each of the 10 classes. However, the core-set baseline selects examples from cat class more than other  as seen in Figure ~\ref{fig:core-set_bar}. A similar trend exists for our approaches too in Figures ~\ref{fig:our1_bar} and ~\ref{fig:our2_bar} where the cat class is selected more.

\section{Conclusion and Future Work}
In this work we proposed a non-iterative and model agnostic method for performing Active Learning using self-supervised features. While being non-iterative our approach is still comparable/slightly worse to the existing state-of-the art approaches. We showed via frequency histogram, the different classes selected in the labelled dataset, and find similar behaviour as to the state-of-the=art approaches.\\

In the future we will like to explore other datasets such as CIFAR-100 and ImageNet~\cite{deng2009imagenet} and also try to further improve our performance with possibly better self-supervised features.

\section{Contributions}
Abhishek implemented the baselines and did experiments with the MoCo-v2 approach.
Shreya did experiments with the SimCLR approach. The project was also advised by Jiaming Song.
\bibliography{ref}
\bibliographystyle{plain}




\end{document}